\documentclass[letterpaper, 10 pt, conference]{ieeeconf}  

\IEEEoverridecommandlockouts                              

\usepackage{graphics} 
\usepackage{mathptmx} 
\usepackage{amsmath} 
\usepackage{amssymb}  
\usepackage{xcolor}
\usepackage{algorithm}
\usepackage[noend]{algpseudocode}
\usepackage{mathtools,amssymb,lipsum,caption}
\usepackage{dirtytalk}
\usepackage{url}

\usepackage{graphicx}
\graphicspath{ {images/} }

\title{\LARGE \bf
Fessonia: a Method for Real-Time Estimation of Human Operator Workload Using Behavioural Entropy
}

\author{Paraskevas Chatzithanos$^{1}$, Grigoris Nikolaou$^{1}$, Rustam Stolkin$^{2}$, Manolis Chiou$^{2}$
\thanks{This work was supported by the UKRI-EPSRC grant EP/R02572X/1 (UK National Centre for Nuclear Robotics).}%
\thanks{$^{1}$University of West Attica, Greece
        {\tt\small parischatz94@gmail.com, nikolaou@uniwa.gr}}%
\thanks{$^{2}$Extreme Robotics Lab and National Centre for Nuclear Robotics, University of Birmingham, UK,
        {\tt\small m.chiou@bham.ac.uk}}%
}

\begin{document}

\maketitle
\thispagestyle{empty}
\pagestyle{empty}


\begin{abstract}
This paper addresses the problem of the human operator cognitive workload estimation while controlling a robot. Being capable of assessing, in real-time, the operator's workload could help prevent calamitous events from occurring. This workload estimation could enable an AI to make informed decisions to assist or advise the operator, in an advanced human-robot interaction framework. We propose a method, named Fessonia, for real-time cognitive workload estimation from multiple parameters of an operator's driving behaviour via the use of behavioural entropy. Fessonia is comprised of: a method to calculate the entropy (i.e. unpredictability) of the operator driving behaviour profile; the Driver Profile Update algorithm which adapts the entropy calculations to the evolving driving profile of individual operators; and a Warning And Indication System that uses workload estimations to issue advice to the operator. Fessonia is evaluated in a robot teleoperation scenario that incorporated cognitively demanding secondary tasks to induce varying degrees of workload. The results demonstrate the ability of Fessonia to estimate different levels of imposed workload. Additionally, it is demonstrated that our approach is able to detect and adapt to the evolving driving profile of the different operators. Lastly, based on data obtained, a decrease in entropy is observed when a warning indication is issued, suggesting a more attentive approach focused on the primary navigation task.


\end{abstract}


\section{Introduction}
Human-Robot Teams (HRT) are becoming increasingly prominent in applications such as autonomous cars \cite{selfdriving}, disaster response \cite{disasterresponse}, and remote inspection \cite{inspection}. In such HRT, humans employ their knowledge and experience about the system, along with information from the user interfaces, to gain awareness of the robot's status. In contrast, the robot's AI agent does not possess the ability to estimate the human's status by default and it should be explicitly provided. Of critical importance w.r.t the human's status, is the level of workload that they experience during task execution. Research shows that using operator workload estimates, has a positive impact on the overall HRT performance \cite{Young2019, Carlson}. Also, in cases where cognitive workload rises, catastrophic consequences could occur to the human involved, to the robot, and even to the entire mission \cite{risk_workload}. Hereupon, agents understanding the state of each other, translates into a more efficient HRT interaction.

This work focuses on cognitive workload (referred to as workload for the rest of the paper) in the context of remotely operated robots, commonly used in safety-critical scenarios such as disaster response. For efficient Human-Robot Interaction (HRI), the robot's AI should not only have awareness of the mission parameters but also be cognizant of the operator's performance and status. Evidence shows a strong correlation between performance and workload \cite{workloadimpact, workloadimpact2}. Thus, methods enabling the estimation of cognitive workload could prove to be particularly important in HRT and especially in variable autonomy systems. For instance, the robot's AI agent could issue warning indications to Human-Initiative systems \cite{Chiou2016IROS_HI}) or adjust its policies to actively alleviate the burden of control in Mixed-Initiative \cite{Chiou2019,gspetou} and shared control systems \cite{pappas2020vfh}). In such cases, behavioral entropy is used as an index of behavioral predictability to characterise operator workload induced by mentally taxing tasks. \cite{Young2019, Goodrich2004}. 

We propose a method for the estimation of the operator's workload in real-time, via the use of behavioural entropy called Fessonia\footnote[3]{Fessonia is the Roman Goddess who aids the weary}. Our method is non-intrusive as it uses the control input of the operator to estimate the workload, an advantage over more intrusive workload estimation methods such as using physiological measures (e.g. EEG \cite{eeg2} and heart rate \cite{workload_survey}). Additionally, compared to post-hoc subjective measures of workload such as the NASA-TLX \cite{Hart1988}, our method offers real-time estimation.

We consider the work of Nakayama et al. \cite{Nakayama1999}, who developed an offline workload estimation system, that utilises steering data to estimate the operator workload. In this work we propose the following novelties. 
First, Fessonia is multidimensional as multiple parameters of the operator's driving profile (e.g. angular and linear velocity commands) can be used.
Second, our method is handling the measurements in real-time resulting in an online estimation of workload.
Third, Fessonia is capable of updating online the operator's driving behavior profile model. We consider the latter to be the main contribution of this paper as it allows the system to adapt its entropy calculation according to the level of the operator's adaptiveness vis-à-vis system and mission familiarity which significantly affect performance.
Dynamically adjusting the robot's understanding of the operator driving profile allows for a more sensitive workload estimation for high performing operators.
Lastly, a Warning And Indication System (WAIS) was developed. This system continuously monitors the workload of the operator and when it exceeds a defined threshold, the system provides an audio-visual indication that suggests a more attentive driving approach. 

\section{Related Work}

Early work utilised the entropy of steering data (angle of steering) and is mostly focused on vehicle driver workload estimation \cite{Boer2000a, Nakayama1999}. Nakayama et al. \cite{Nakayama1999} presented a methodology based on the assumption that the driver's steering behaviour tends to become more unpredictable, under heavy workload. 
They proposed a metric based on the steering angle and steering angle estimations, to formulate a methodology that detects unpredictable behaviour using behavioural entropy and thus quantify the driver's workload.
In \cite{Boer2000a}, they used a similar methodology with \cite{Nakayama1999}, but they substituted the Taylor expansion approximation with a second-order auto-regressive model. This method is more sensitive to roads with high curvatures or roads with sudden turns.

Other related work is focused on virtual reality systems. In \cite{Reinhardt2019} they presented a methodology that combines sample entropy and template matching, applied in a virtual reality system. Their proposed methodology allows for the creation of different input templates, and the entropy is calculated based on the number of similar templates. Even though this method is computationally expensive, it provides a relative noise-free multi-dimensional metric for workload estimation.

One metric that can be used in real-time is presented in \cite{Young2019} and is called Discrete N-Dimensional Entropy of behaviour (DNDEB). The discretized inputs are used in combination with a long short-term memory network to produce confusion matrices for a predefined data window, and calculate entropy based on those confusion matrices. The advantages of this work are that the metric accepts discrete inputs of any size, and is tunable to the specific application. In \cite{Young2019} the main focus was to use entropy as a means to estimate the performance of human-machine rather than operator workload.

Our research shares similarities with approaches that employ behavioural entropy to estimate the workload of vehicle drivers \cite{Nakayama1999,Boer2000a}, while further being motivated by works that appraise human workload in virtual reality \cite{Reinhardt2019} and human-robot team performance in robotic wheelchairs \cite{Young2019}. To the best of the authors' knowledge, most related work focuses on using entropy as a post-hoc measure of human workload, except for \cite{Young2019} where they used entropy to develop a collision prediction mechanism. Furthermore, previous research assumes that after the initial training of the operator, the levels of familiarity, experience, and driving ability to stay static \cite{Boer2000a,Young2019}. In this work, we aim to address these issues by proposing a framework for real-time workload estimation that takes into consideration the operator's current condition and evolving driving profile. Furthermore, it uses said estimates on a warning indication system that issues warnings during elevated operator workload.



\section{Problem Formulation and Development}\label{form}

\begin{table*}
\centering
\begin{align}\label{eqn:taylor}
\begin{bmatrix}
    E_{l}(n)\\
    E_{a}(n)
\end{bmatrix} = 
\begin{bmatrix}
    E_{l}(n-1)+\big(E_{l}(n-1)-E_{l}(n-2)\big)+\bigg(\big(E_{l}(n-1)-E_{l}(n-2)\big)+\big(E_{l}(n-2)-E_{l}(n-3)\big)\bigg) \\
    E_{a}(n-1)+\big(E_{a}(n-1)-E_{a}(n-2)\big)+\bigg(\big(E_{a}(n-1)- E_{a}(n-2)\big)+\big(E_{a}(n-2)-\textbf{}(n-3)\big)\bigg)
\end{bmatrix}
\end{align}

\end{table*}

We consider the problem of estimating the workload of an operator, controlling a remotely situated robot and conducting a mentally demanding task. For example, a search and rescue scenario where a human-robot team is assigned to navigate and explore an environment with the objective to detect victims. These situations are very taxing and stressful to human operators as they often require multitasking on top of the cognitively demanding robot control \cite{search}.

Similar to \cite{Nakayama1999}, we make an informed assumption that high operator workload correlates with unpredictable driving behaviour profile. While the cognitive load imposed increases, the operator is likely to make errors during the primary task (i.e. robot navigation), and thus is more prone to resort to abrupt error corrective commands and more specifically jerky velocity commands of the joypad control, i.e. linear command velocity $Cmd\_Vel_{l}$ and angular command velocity $Cmd\_Vel_{a}$. 

To detect these jerking motions we use a second-order Taylor expansion model (Eq. \ref{eqn:taylor}) that estimates a typical expected operator workload profile (referred to as typical operator workload), using operator command parameters such as linear $E_{l}$ and angular $E_{a}$ velocity commands. We used behavioural entropy $Hp$ in the estimation errors since it is a commonly used measure of unpredictability, i.e. high entropy reflects high workload.


\subsection{Behavioural Entropy With Multiple Dimensions}
Initially, we defined the parameters that are required to calculate behavioural entropy, which are the angular and linear velocity commands of the operator (Eq. (\ref{eqn:inputs})).
\begin{equation}\label{eqn:inputs}
Measurement Input: 
\begin{bmatrix}
    Cmd\_Vel_{l}(n)\\
    Cmd\_Vel_{a}(n)
\end{bmatrix}
\end{equation}
These commands were sampled at a rate of 20 Hz and a three sample moving average filter was used, in order to reduce measurement noise. The frequency of filtered samples were roughly 7 Hz, which is equal to the humans manual tracking delay \cite{McDonnell1966, sheridan1974man}.
We used a second-order Taylor expansion to model the typical driving profile at time-step n (Eq. \ref{eqn:taylor}). Let $E_{l}(n)$ and $E{a}(n)$ be the estimated linear command velocity and the estimated angular command velocity of the robot at time-step n respectively. Using the model from Eq. \ref{eqn:taylor}, the estimation errors were calculated:
\begin{equation}\label{eqn:esterrors}
    \begin{bmatrix}
        Estim\_Error_{l}(n)\\
        Estim\_Error_{a}(n)
    \end{bmatrix} = 
    \begin{bmatrix}
        Cmd\_Vel_{l}(n)\\
        Cmd\_Vel_{a}(n)
    \end{bmatrix} - 
    \begin{bmatrix}
        E_{l}(n)\\
        E_{a}(n)
    \end{bmatrix}
\end{equation}

\begin{figure}[]
\centering
\includegraphics[scale=0.4]{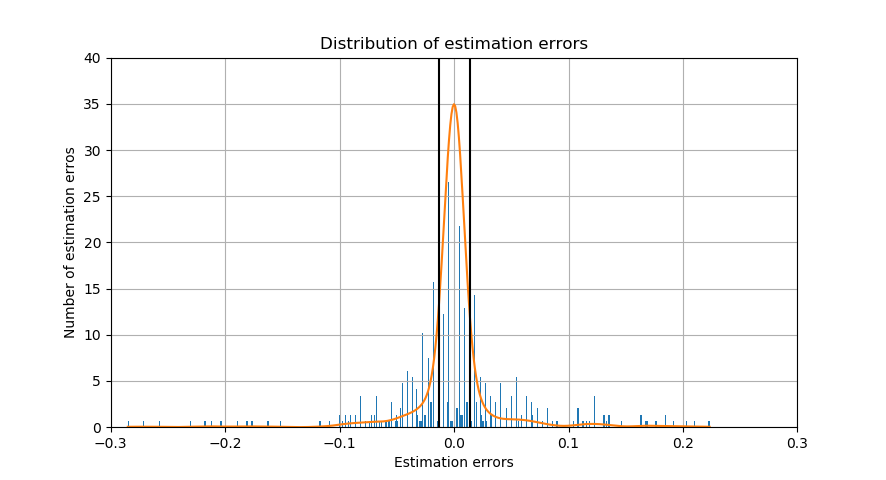}
\caption{Example distribution of estimation errors ($ E_{a}$) from the experiments. The two vertical black lines represent the 90\textsuperscript{th} percentile value $\alpha$.}
\label{example_distr}
\end{figure}

\begin{figure}[]
\centering
\includegraphics[scale=0.45]{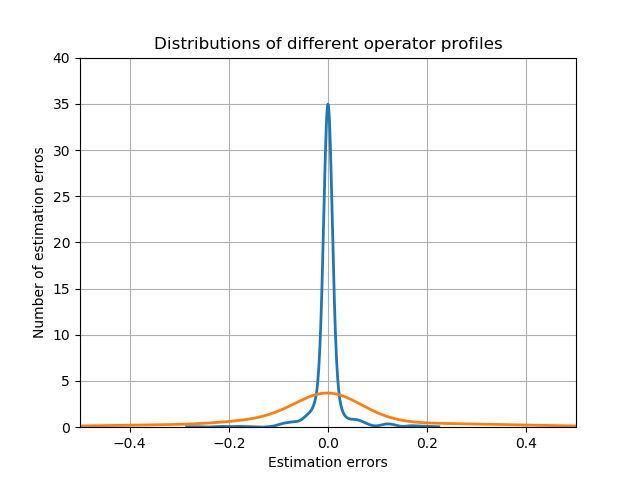}
\caption{Frequency distribution of angular velocity's estimation errors. The \textcolor{blue}{blue distribution} is derived from an experiment where no workload was imposed on the operator, while the \textcolor{orange}{orange distribution} is derived by an operator with very high workload.}
\label{frequency distribution}
\end{figure}

For calculating the entropy, the 90\textsuperscript{th} percentile (value $\alpha$) of the estimation error frequency distribution was calculated. One example of the distribution of estimation errors from our trials can be seen in Fig. \ref{example_distr}. The value $\alpha$ is indicative of the different operator driving profiles and ultimately operator workload. When an operator deviates from typical behaviour, the prediction errors recede from zero, making the estimation error distribution wider (Fig. \ref{frequency distribution}) and $\alpha$ bigger.

Using the value $\alpha$ of the distribution, we divided the distribution into 9 bins according to \cite{Khinchin}, resulting in 8 bin boundaries $[-5\alpha, -2.5\alpha, -\alpha, -0.5\alpha ,0.5\alpha, \alpha, 2.5\alpha, +5\alpha]$. 
The operators need to first undergo a trial-run session (see \ref{s}), in which they operate the robot without any specific mission task apart from driving, to calculate the initial baseline driving behaviour profile. 
We use the entropy \cite{Khinchin}, defined as:
\begin{equation}\label{eqn:entropy}
Hp = \sum_{i=1}^{9} -p_i \times Log_9(p_i)    
\end{equation}
\begin{equation}\label{eqn:totalentropy}
Total\ Entropy : 
\begin{bmatrix}
    Hp_{l}\ 
    Hp_{a}
\end{bmatrix} \times 
\begin{bmatrix}
    0.5 \\
    0.5
\end{bmatrix}
\end{equation}
The values $p_i$ denote the frequency of the errors that fall in the respective bins while $i$ denotes the number of bins.
In our implementation the entropy calculation is two dimensional $(Hp_{l}, Hp_{a})$, as the input has one linear ($ Cmd\_Vel_{l}$) and one angular ($Cmd\_Vel_{a}$) velocity command dimension. The proposed method has the additional advantage of being designed for modularity. This can be attributed to the fact that it can be trivially augmented with the inclusion of more inputs and thus more dimensions. Here, the two dimensions of the entropy are considered equally significant to the estimated Total Entropy (Eq. \ref{eqn:entropy}). 


The estimation errors are calculated every 0.15 seconds however, the entropy calculation occurs periodically with a defined frequency.
We empirically observed the average corrective manoeuvre span to be less than .25 seconds and chose entropy computation period $10\times 0.25 = 2.5 seconds$ (0.4 Hz). Entropy calculation less than 0.2 Hz or more than 0.4 Hz results in a slow, or unreliable workload estimation respectively. This frequency also affects the batch size of estimation errors included in the entropy computation. To elaborate, when the entropy is calculated every 3 seconds, the estimation errors handled are $\frac{3}{0.15}=20$. 


\subsection{Driver Profile Update}

In this work, we explicitly consider the: adaptation of humans to different situations; individual differences in task-related skills between operators; evolvement of task-related skills over time. It is natural for an operator to be constantly improving their performance by experience and having the ability to adapt to dynamic environments.
A static operator profile will produce estimation errors that falsely indicate low workload when high workload is imposed, as these errors lie in closer to zero bins, rather than further from zero bins. 

In our work, we developed an algorithm called Driver Profile Update (DPU) that monitors the entropy history and updates the operator profile characteristics ($\alpha$ and bins) accordingly. If no prior knowledge is available, the initial thresholds (i.e. average and std variables) can be defined as extremely high values. We used entropy history acquired from a familiarization session (Section \ref{s}). If the average and standard deviation of the 100 previous entropy history time-steps become lower than the initial thresholds, DPU saves the last 100 estimation errors to the error history and utilises said error history to calculate an updated value for $\alpha$ and new bins (see Algorithm \ref{alg:dist_update}). The defined thresholds are also updated to the newly updated average and standard deviation of the last 100 entropy history.
\begin{algorithm}  \label{algorithm}
\caption{Driver Profile Update (DPU)}\label{alg:dist_update}
\begin{algorithmic}[]
\Require \State entropy\_history, error\_history, estimation\_errors[-100:], \\ average = mean(entropy\_history), \\ std = std(entropy\_history)
\end{algorithmic}
\begin{algorithmic}[1]
\Procedure{Profile\_Update}{}
    \If{$mean(entropy\_history[-100:])<average$}
        \If{$std(entropy\_history[-100:])<std$}
            \State $error\_history \leftarrow estimation\_errors[-100:]$
            \State $average = mean(entropy\_history[-100:])$
            \State $std = std(entropy\_history[-100:])$
            \State\textbf{return} $percentile\_\alpha(error\_history), new\_bins$ 
        \EndIf
    \EndIf
\EndProcedure
\end{algorithmic}
\end{algorithm}

\subsection{Warning And Indication System (WAIS)}
To assist the operator in their mission, a Warning And Indication System (WAIS) is proposed. WAIS continuously monitors the entropy calculations and issues warning indications, according to the entropy produced by the operator's manoeuvres. When the operator manoeuvres through a challenging turn, they might momentarily experience high workload. To alleviate the effect of those sudden entropy \say{spikes}, we use a moving average filter of five computations.

WAIS informs the operator about their workload with two indications, one for normal and one for erratic operation (see Fig. \ref{sim}). 
The operator sees a \say{NORMAL OPERATION} indication until the moving average is greater than a defined threshold, which was empirically assigned at 0.6 to capture severe workload escalations. We concluded to this threshold empirically by monitoring the entropy levels during no-workload test runs, where the entropy spikes did not exceed 0.6.
An audio-visual warning message is issued on the GUI, to alert of a possible detrimental operation. The audio warning is a soft "ping" sound to attract the attention of the operator. The visual warning is a \say{HIGH WORKLOAD} indication that substitutes the "NORMAL OPERATION" indication in the GUI.

\subsection{Method overview}
\begin{figure}[]
\centering
\includegraphics[scale=0.28]{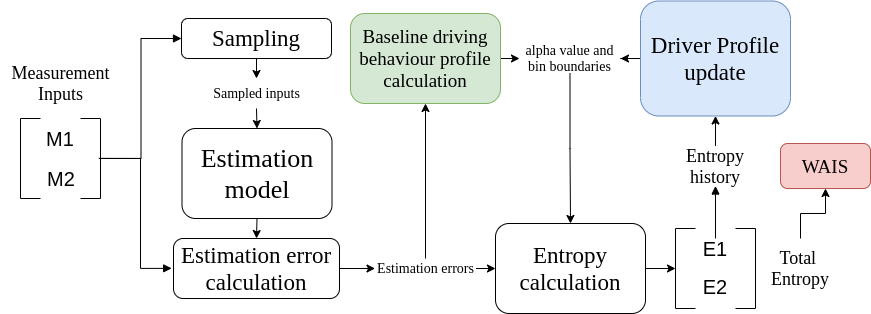}
\definecolor{Mycolor1}{HTML}{B5E0B2}
\definecolor{Mycolor2}{HTML}{AFD5E5}
\definecolor{Mycolor3}{HTML}{ffcccc}

\caption{Diagram of Fessonia. The initial \colorbox{Mycolor1}{baseline driving} \colorbox{Mycolor1}{behaviour profile calculation} computes the 90\textsuperscript{th} percentile ($\alpha$ value) of the error distribution after the  operator's trial-run session, while the \colorbox{Mycolor2}{DPU} continuously monitors the entropy output to sample more estimation errors suitable for updating the estimation error distribution (value $\alpha$). \colorbox{Mycolor3}{WAIS} issues warnings based on operator workload.}
\label{my_block_diagram}
\end{figure}

The above-presented methods are handled by three subsystems, namely the asynchronous entropy calculation (including sampling, estimation model, estimation error calculation and initial baseline driving behaviour profile calculation), the DPU, and the WAIS (see Fig. \ref{my_block_diagram}). What follows is the testing and evaluation of the above subsystems.

\section{Experimental study}
We conducted a preliminary experimental evaluation of the proposed methods on a mission inspired by disaster response and remote inspection scenarios, where an operator controlling a remote robot, is tasked with exploring an environment (e.g. for points of interest such as people trapped in debris). The experiments aimed to evaluate the: a) accuracy of the workload estimation and how it quantifies different levels of workload; b) ability of the method to detect the adapting driving profile of human operators; c) effect that the warning indication system has on behavioural entropy and as a consequence on the workload of an operator.

\subsection{Apparatus and Software}

All software was developed in the Robot Operating System (ROS) and is available in our code repository\footnote[4]{The code for the experiments is provided under MIT license in the Extreme Robotics Lab GitHub repository: \url{https://github.com/uob-erl/hrt\_entropy}}. The environment and the robot were simulated in Gazebo. The simulated robot was a Husky model equipped with a camera to provide a video feed and a 2D LiDAR laser sensor for localization. The specific experimental setup based on simulation was selected because repeatability was considered of high importance.
Moreover, during the operation of remote robots, the interface/cockpit and the operator's experience is almost identical in simulation of this work as it is in reality.  
As Gazebo is a high-fidelity robotic simulator the realism or meaningfulness of the results were not compromised.

\begin{figure}[]
\centering
\includegraphics[scale=0.13]{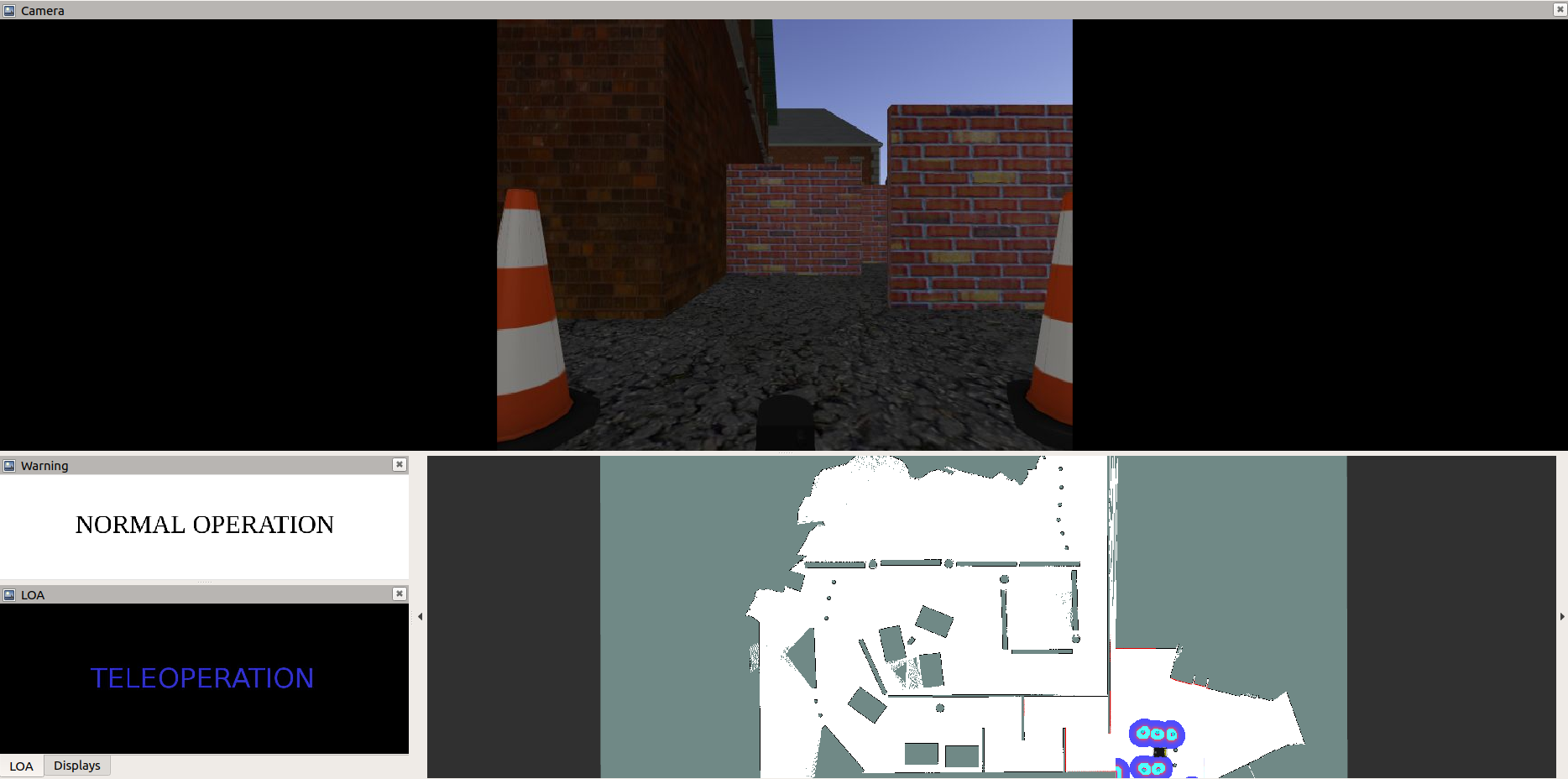}
\caption{The Graphical User Interface (GUI) shows the position of the robot relative to the map, the operation status of the robot, video streaming from the robot's on-board camera, and a warning indication (WAIS).}
\label{sim}
\end{figure}

Participants operated the robot via a joypad while acquiring situation awareness (SA) via a GUI (see in Fig. \ref{sim}). The GUI provides information about the position of the robot relative to the map, an operation status indication (shows if the robot is stopped or teleoperated), a workload warning indication, and video streaming from the robot's onboard camera. Moreover, the GUI has a secondary window for the high workload secondary task (described below) used to evaluate the impact of workload on entropy. 
Finally, the operator used a computer mouse to execute the secondary tasks that are designed to assess the workload estimation (see Section \ref{s}).

\subsection{Experimental Protocol} \label{s}
\begin{figure}[]
\centering
\includegraphics[scale=0.25]{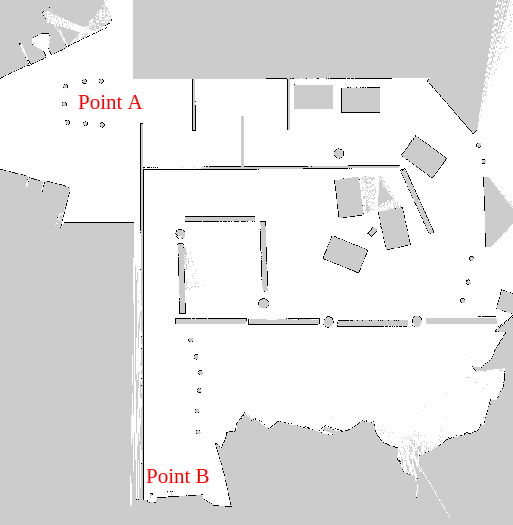}
\caption{Simulation arena map. Points A and B are points of interest. The operator has to navigate from point A to point B and back four times for each experiment.}
\label{map}
\end{figure}

Three participants took part in this study. Two of the participants had previous experience with similar control configurations and interfaces while the third one was a complete novice. Before the experiment, participants were trained by driving around the robot arena for five minutes to get familiar with the system and accustomed to the controls (see Fig. \ref{map}). After said familiarization, the participants navigated the arena, without any secondary tasks, for 10 minutes to develop their initial baseline driving behaviour profile (i.e. the value of $\alpha$). 
Each participant performed 3 experimental trials. In each trial, the participants were required to navigate the robot from point A to point B of the map (Fig. \ref{map}) four times, except from the first experiment trial which included 12 round trips from A to B. Between each experiment, the participants took a 15-minute break to minimize their fatigue levels, which could affect the level of workload estimates. The instruction given to the participants was to navigate the robot at a comfortable pace while avoiding collisions. 

In the first trial, participants were tasked to operate the robot focusing only on the navigation task. This experiment was conducted to evaluate the DPU system.
In the second experiment, the workload estimation algorithm was exclusively evaluated. To validate the impact that workload has on entropy, three secondary tasks were introduced in sequence, from higher to lower workload impact.
The cognitive impact the secondary tasks inflict might vary for each participant. To to be consistent and normalise the impact across all participants, we gave specific instruction to perform the secondary tasks at a comfortable pace with no time restrictions.
The following tasks were inspired by previously validated secondary tasks from the literature (\cite{Nakayama1999} \cite{Boer2000a}). The High workload task includes pressing one out of four buttons, upon request on a secondary GUI window. The secondary GUI will play a sound and request the press of a specific button. The Medium workload task is counting down from 500 by subtracting 3. Finally, during the Low workload task the maximum speed of the robot was doubled and as a result, the robot became more sensitive to small control commands.
The high workload secondary task proposed here simulates operating a touch panel tested in \cite{Nakayama1999} or the press of a button from \cite{Boer2000a}, and the medium workload secondary task is similar to the mental arithmetics from \cite{Nakayama1999,Boer2000a}. 

In the last experiment, the participants were required to navigate the arena, while the high workload imposing secondary task was in effect. 
The instructions were the same as before but when an indication was issued, participants were instructed to ignore completely the secondary task and continue conducting the primary task (operating the robot), until the warning indication ceased. This experiment was designed to evaluate the effect WAIS has on the operator workload. Across all experiments, the focus was to investigate the effect of the secondary tasks on the operators' workload rather than performance on tasks.

\begin{figure}[]
\centering
\includegraphics[scale=0.45]{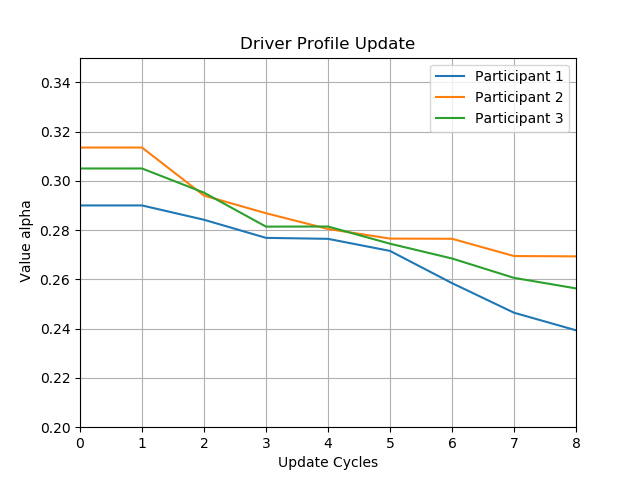}
\caption{DPU system results represent operators' evolving driving. The three participants had already conducted a trial-run session to compute their initial driver behaviour profile. If the necessary conditions (Algorithm \ref{alg:dist_update}) are met, the bins and value of $\alpha$ are updated. In a half-hour trial, $\alpha$ has been reduced for all participants, meaning participants' ability to drive has improved.}
\label{dpu}
\end{figure}

\section{Results}

\begin{figure}[]
\centering
\includegraphics[scale=0.45]{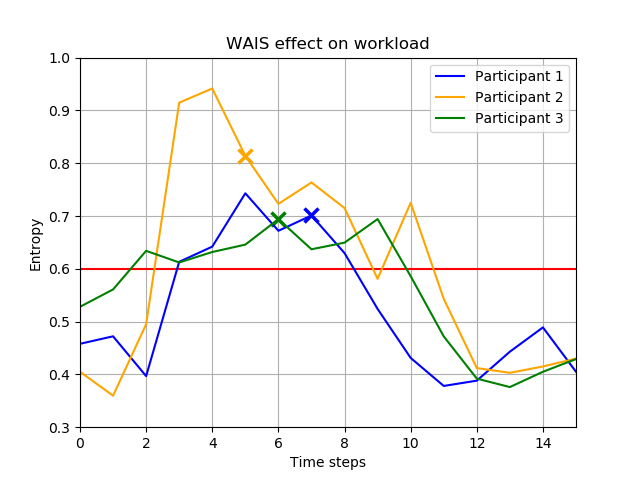}
\caption{WAIS effect on entropy during an exemplary 37.5 second time window taken for each participant's third 45 minute experimental trial. The \textcolor{red}{red} line is the empirically defined threshold for Normal Operation and the crosses are where the WAIS issued a warning. 
Participant 1 immediately listens to the indication and recovers in approximately 5 seconds, while Participants 2 and 3 had a slower response to the indication, roughly 17 and 10 seconds.}
\label{indication}
\end{figure}


\begin{table*}[]
\centering
\caption{Entropy descriptive statistics for three participants on three sequentially increasing workload imposing tasks}
\label{table:entropy_results}
\begin{tabular}{l|cccccccc}
\begin{tabular}[c]{@{}l@{}}Participants and\\ secondary task\\ trial portions\end{tabular} & \multicolumn{2}{c}{Baseline trial portion} & \multicolumn{2}{c}{Low workload trial portion} & \multicolumn{2}{c}{Medium workload trial portion} & \multicolumn{2}{c}{High workload trial portion}  \\
                                                                                           & \textit{M} & \textit{SD}                   & \textit{M} & \textit{SD}                       & \textit{M} & \textit{SD}                          & \textit{M} & \textit{SD}                         \\ 
\hline
Participant 1                                                                              & 0.1375     & 0.1298                        & 0.1946     & 0.1204                            & 0.3493     & 0.1184                               & 0.4441     & 0.1051                              \\
Participant 2                                                                              & 0.1641     & 0.1319                        & 0.2238     & 0.1279                            & 0.3501     & 0.1201                               & 0.4918     & 0.1174                              \\
Participant 3                                                                              & 0.1407     & 0.1346                        & 0.1993     & 0.1308                            & 0.3418     & 0.1247                               & 0.4248     & 0.1091                              \\ 
\hline
\textbf{Average}                                                                           & 0.1474     & 0.1321                        & 0.2059     & 0.1264                            & 0.3471     & 0.1211                               & 0.4546     & 0.1105                              \\
\hline
\end{tabular}
\end{table*}

In the first trial, results show that the value $\alpha$ (operator driving profile) is decreasing over the duration of the experiment, suggesting a predictable behaviour (Fig. \ref{dpu}).
The second trial showed that the multi-dimensional entropy was proportional to the different levels of imposed workload (see Table \ref{table:entropy_results}). Finally, from the third trial, three examples of issued warnings from the third 45-minute trial are presented for each operator (Fig. \ref{indication}). These examples show a duration of an average 10 seconds, to decrease the entropy after a warning was issued.


\section{Discussion, Limitations and Future Work}
Evidence from the preliminary evaluation suggests that the proposed method estimates successfully the workload using multiple control command parameters given by the operator. The descriptive statistics presented in Table \ref{table:entropy_results} show a considerable distinction between the secondary tasks with different workload impacts. This is in accordance with results presented in the literature \cite{Boer2000a, Nakayama1999, Young2019, Reinhardt2019} which show that entropy can capture different levels of workload. 
Due to the real-time functionality of the method, the entropy calculation, in some situations, could be zero. For instance, where the operator follows a typical driving profile and thus is predictable, all estimation errors will fall in the bin around zero. In other words, the rate of errors will be 1 for one specific bin, resulting in an entropy estimation equal to $Hp = 1Log^9(1)=0$, and in consequence in greater standard deviation. Nonetheless, the standard deviation does not have a substantial effect on the workload estimation. Results suggest that the zero entropy might slightly increase the standard deviation, but the standard deviation is a stronger indication of individual driving differences among the participants.

The DPU system results support our initial assumption, which is that the operators can adapt to certain situations and control schemes, in addition to becoming more competent in the task over time. 
The $\alpha$ characterises the driving profile of the operator. Smaller $\alpha$ would suggest a more predictable driving profile  (i.e. typical operation), and over time, decreasing $\alpha$ suggests that the operator becomes more predictable. One can argue that operators who are becoming more predictable are essentially adapting their driving profile, becoming more familiar with the task, and increasing the overall performance of the HRT. 
Every participant showed a different adapting process, but all of them adapted on some level. In addition, each experiment was conducted for 30 minutes and $\alpha$ did not fully converge for any of the participants. This means that they could get even more acclimated to the situation. The results also suggest that the initial trial-run session to calculate the value $\alpha$ is redundant (which was required in previous literature \cite{Boer2000a,Nakayama1999}). The system can start with a relatively high $\alpha$ value and the system can adjust this value due to the DPU if the operator consistently produces lower entropy computations. 

The WAIS trials showed promising results. In the experiments, the participants were instructed to completely ignore the secondary task when a warning was issued. The results present a 21.3\% average decrease in entropy across all participants. Specifically, Participant one had a 21\% decrease in $\alpha$, Participant 2 had a 14\% in $\alpha$ and Participant 3 had an 18\% decrease in $\alpha$ value. The entropy reduction was more than expected since the main instruction was to only focus on the navigation task. In a real-world scenarios the operator could be multitasking during a mission and thus a warning could make them aware of critical performance and safety impairment factors. Drawing the operator's attention aims to remind them to focus on the safe driving of the robot to avoid potential accidents. Depending on the mission, the existence of a specific protocol to be followed in the case of a warning will be beneficial. Additionally, WAIS displays the sensitivity of the entropy computation to driving behaviours. The participants become aware of the deteriorated performance upon indication and change their driving behaviour which is encapsulated by the reduced entropy.

Assuming that all control command dimensions equally contribute to the total entropy, might not hold true for different applications and environments. In future work, a dynamic thresholding system could assign thresholds proportionally to the correlation coefficient of each control command dimension to entropy.
Despite the promising results of WAIS, in a real high-stress scenario, a warning system could potentially increase the workload and stress levels of an operator if not integrated carefully with the mission protocols. More research is needed to appropriately design WAIS, to minimize the stress and workload imposed by the indication itself, while preserving the value of the warning itself. 
Lastly, high entropy estimations could be the result of increased operator fatigue over time, rather than increased workload. Our experimental design aimed to minimize this possibility. However, in future work, to differentiate between fatigue and workload, measurements from other non-intrusive methods (e.g. eye-tracking etc) can be used. In literature there have been examples of using eye tracker systems to detect fatigue levels \cite{eyetracker}. Our system could benefit from measurements of such detectors.

\section{Conclusion}

In this study, we propose Fessonia, a method for estimating the cognitive workload of an operator controlling a robot, via the use of behavioural entropy. This method's contribution is the capability of estimating operator workload given multiple parameters of the operator’s driving profile. An additional contribution is the Driver Profile Update (DPU) algorithm which continuously observes the performance history of the operator and adapts the workload estimation based on the adapted operator profile. Lastly, a warning indication system (WAIS) is presented, that advises the operator based on workload estimates.
Fessonia was evaluated in a navigation scenario. The results indicate that our proposed method captures and successfully estimates different levels of workload. Furthermore, our proposed DPU algorithm can detect individual driving profiles, improvements in operator performance, and adapts accordingly. Hence, the workload estimation becomes more responsive to the level of competence of the operator. Lastly, the use of WAIS resulted in an average 25\% entropy reduction across all participants. The future work's focal point is to better improve our method by redesigning WAIS to minimize any potential stress and workload imposed by the suggestion itself. Furthermore, it is under consideration to include additional dimensions, that would provide Fessonia with the ability to distinguish between operator fatigue and imposed workload. Lastly, the promising experimental results, enable the future integration of our method in variable autonomy human-robot teams. Said system could utilise these entropy calculations to actively support the human operator, e.g. by completely relinquishing control from the operator over to the robot's AI agent.

\addtolength{\textheight}{-1cm}
\bibliographystyle{IEEEtran}
\bibliography{root}
\end{document}